\documentclass[sigconf,nonacm]{acmart}

\usepackage{array}
\usepackage{amsmath}
\usepackage{booktabs}
\usepackage{makecell}
\usepackage{multirow}
\usepackage{tabularx}

\setcopyright{none}
\renewcommand\footnotetextcopyrightpermission[1]{}

\AtBeginDocument{%
  }

\begin{document}

\title[RAPID for Interpretable Disaster Damage Assessment]{RAPID: A Reproducible Multi-Agent Pipeline for Interpretable Disaster Damage Assessment from Satellite and Street-View Imagery}

\author{Yifan Yang}
\email{yyang295@tamu.edu}
\affiliation{%
  \institution{Texas A\&M University}
  \city{College Station}
  \state{Texas}
  \country{USA}}

\author{Wenjing Gong}
\email{wenjinggong@tamu.edu}
\affiliation{%
  \institution{Texas A\&M University}
  \city{College Station}
  \state{Texas}
  \country{USA}}

\author{Kaili Zhang}
\email{kelly_zhang@tamu.edu}
\affiliation{%
  \institution{Texas A\&M University}
  \city{College Station}
  \state{Texas}
  \country{USA}}

\author{Lei Zou}
\authornote{Corresponding author.}
\email{lzou@tamu.edu}
\affiliation{%
  \institution{Texas A\&M University}
  \city{College Station}
  \state{Texas}
  \country{USA}}

\author{Zhengzhong Tu}
\email{tzz@tamu.edu}
\affiliation{%
  \institution{Texas A\&M University}
  \city{College Station}
  \state{Texas}
  \country{USA}}

\author{Hao Li}
\email{hao.li@nus.edu.sg}
\affiliation{%
  \institution{National University of Singapore}
  \city{Singapore}
  \country{Singapore}}

\author{Zongrong Li}
\email{zongrong@tamu.edu}
\affiliation{%
  \institution{Texas A\&M University}
  \city{College Station}
  \state{Texas}
  \country{USA}}

\author{Xinyue Ye}
\email{xye10@ua.edu}
\affiliation{%
  \institution{The University of Alabama}
  \city{Tuscaloosa}
  \state{Alabama}
  \country{USA}}

\begin{abstract}
Due to the increasing frequency and intensity of extreme climate events, there is a clear demand for intelligent, scalable, and autonomous approaches to disaster damage assessment. Existing methods, largely based on supervised learning and task-specific fine-tuning strategies, struggle to generalize under domain shifts, long-tailed data distributions, and heterogeneous geospatial data sources, especially in disaster scenarios. In addition, they often lack the ability to integrate and reason effectively across multimodal geospatial information, such as satellite images and street-view images. In this paper, we introduce RAPID, a reproducible multi-agent pipeline for interpretable disaster damage assessment — encompassing the assessment of damage levels, interpretation of damage types and degrees, and generation of actionable suggestions for response, remediation, and recovery. RAPID coordinates multiple specialized agents to perform cross-view understanding, image restoration, structured damage recognition, and geographical reasoning across heterogeneous data modalities. Without task-specific fine-tuning, RAPID supports zero-shot damage assessment by jointly leveraging complementary information from remote sensing and ground-level perspectives. Furthermore, the system produces fine-grained, interpretable damage assessments and automatically generates location-specific, decision-relevant disaster reports to support early-stage emergency response. We evaluated and demonstrated the performance of the proposed RAPID framework across multiple disaster scenarios, including hurricanes, floods, wildfires, and earthquakes, using different sets of cross-view imagery inputs, including street-view images before and after the disaster, remote sensing imagery after the disaster, and street-view image pairs. Experiments show that RAPID achieves an overall accuracy of 0.92 for multi-disaster-type classification and up to 0.627 for cross-view damage severity prediction, highlighting its potential as a foundational framework for autonomous disaster intelligence.
\end{abstract}



\maketitle

\section{Introduction}
Against the backdrop of increasingly frequent extreme natural disasters, achieving a rapid, accurate, and spatially specific damage assessment is crucial for emergency response, resource allocation, and post-disaster recovery\cite{Kerle2024DisastersRiskAssessment,Khan2023SystematicReviewDisaster,Kirpalani2024TechnologyDrivenApproachesEnhance,Yu2018BigDataNatural}. In recent years, deep learning and remote sensing techniques have become the dominant paradigm for disaster damage assessment, with convolutional and transformer-based models trained on satellite or aerial imagery to classify and quantify structural damage \cite{Zou2023GeoAIDisasterResponse}. Large vision-language models (VLMs) and large language models (LLMs) have been introduced to interpret multi-source disaster imagery and generate readable damage descriptions \cite{Chen2024IntegrationLargeVision, Lei2025HarnessingLargeLanguagea}.

However, existing methods face several key limitations. First, most techniques rely on fine-tuning pre-trained models and require extensive manual annotation, which is difficult to implement in time-sensitive disaster scenarios \cite{yang2026damagearbiter}. Second, many models are built on datasets from single events, resulting in limited generalization capabilities across different regions and disaster types \cite{Zou2023GeoAIDisasterResponse} \cite{Yang2025Hyperlocaldisasterdamagea}. Third, traditional research often relies on single-modal data, such as only satellite imagery or street-view imagery, neglecting the complementarity of multi-source information in capturing the diverse types and intensities of disaster impacts in real-world scenarios \cite{Ahn2025Generalizabledisasterdamage, Chen2025Brightgloballydistributed, Chen2024IntegrationLargeVision, Lei2025HarnessingLargeLanguagea, Ma2025MultimodalMultilingualMultidimensionalb}.

Recent developments in geospatial artificial intelligence (GeoAI) and autonomous agent technologies have provided new avenues for addressing the aforementioned bottlenecks \cite{Bell2025EarthAIUnlockinga,Yin2025LLMenhanceddisastergeolocalizationa,Zuo20254KAgentAgenticAnya}. Benefiting from large-scale pre-training, GeoAI foundation models can perform zero-shot or few-shot analysis with reduced task-specific retraining, while their multimodal, cross-modal alignment further enables joint reasoning across heterogeneous disaster-related data, including satellite imagery, street-view imagery, text descriptions, and social media information \cite{Bell2025EarthAIUnlockinga,yang2024geolocator,Yang2025Hyperlocaldisasterdamagea,Yin2025LLMenhanceddisastergeolocalizationa}. Building on such models, autonomous agents can be designed to support task planning, multi-source data utilization, model result arbitration, and real-time interpretable output, shifting disaster assessment from traditional pattern recognition toward reasoning-based autonomous understanding \cite{Akinboyewa2025GISCopilotautonomousa,Durante2024AgentAISurveying,Li2025Crossviewgeolocalizationdisaster}.

This study introduces RAPID, a \textbf{R}eproducible multi-\textbf{A}gent \textbf{P}ipeline designed for cross-view \textbf{I}ntelligence in disaster \textbf{D}amage assessment. Instead of focusing on a single predictive task, RAPID aims to enable a detailed, interpretable understanding of disasters from cross-view geospatial observations, including satellite imagery and street-level imagery. Specifically, this study is guided by three research questions (RQ).

\begin{itemize}
    \item \textbf{RQ1}: How can an autonomous multi-agent framework be designed to support zero-shot damage assessment using cross-view satellite and street-view imagery?
    \item \textbf{RQ2}: How well does the proposed framework support accurate, interpretable, and generalizable damage assessment across diverse disaster scenarios?
    \item \textbf{RQ3}: What are the strengths, uncertainties, and practical limitations of applying this agentic cross-view framework for real-world disaster management decision support?
\end{itemize}

RAPID is an autonomous multi-agent framework composed of four collaborative agents: a disaster perception agent, an image restoration agent, a damage recognition agent, and a disaster reasoning agent. This framework forms an end-to-end workflow that connects scene understanding, image enhancement, structured damage assessment, and high-level semantic reasoning. This agentic design directly addresses our three research questions: it specifies how a cross-view multi-agent pipeline can be constructed for zero-shot damage assessment (RQ1); by coordinating specialized agents rather than relying on a single task-specific model, it supports accurate, interpretable, and generalizable assessment across diverse disaster scenarios (RQ2); and its modular, reasoning-based structure makes the strengths, uncertainties, and practical limitations of the framework explicit for real-world decision support (RQ3). 

\begin{figure*}[tbp]
  \centering
  \includegraphics[width=0.75\textwidth]{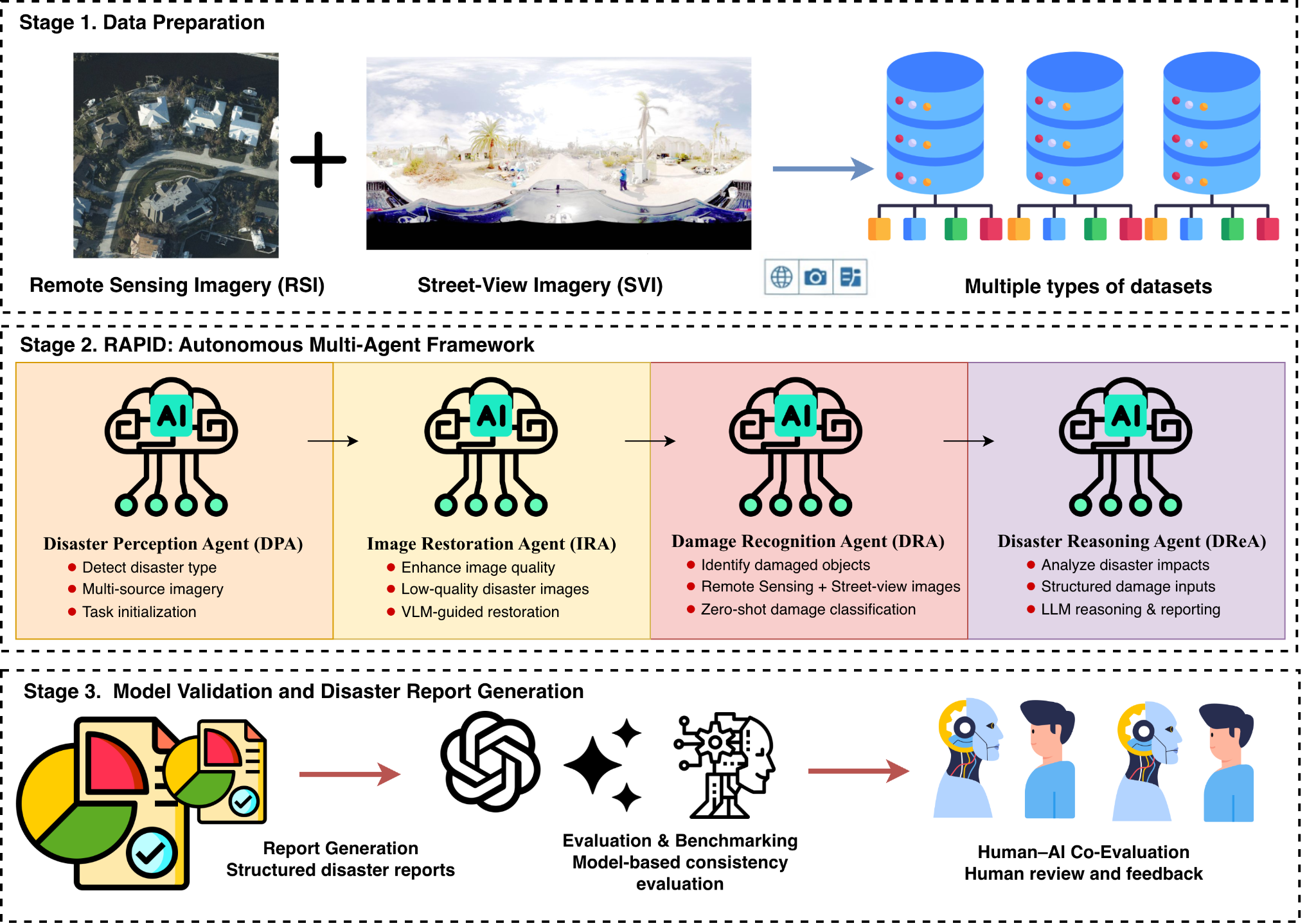}
  \caption{RAPID: an autonomous multi-agent framework for disaster damage assessment.}
  \Description{Pipeline diagram showing cross-view disaster data flowing through disaster perception, image restoration, damage recognition, and disaster reasoning agents.}
  \label{fig:framework}
\end{figure*}

\section{Related Work}
\subsection{Image Analysis for Disaster Damage Assessment}

Disaster damage assessment has advanced rapidly through the analysis of disaster imagery, ranging from top-down remote sensing to ground-level street views, supported by pre-trained models and large-scale benchmark datasets. While overhead satellite imagery offers broad spatial coverage, it often struggles to resolve facade-level structural damage, motivating the use of complementary ground-level observations. Recent studies have investigated how different viewpoints, temporal information, and learning architectures can jointly improve the accuracy and interpretability of post-disaster analysis. Yang et al.~\cite{Yang2025Hyperlocaldisasterdamagea} demonstrated that bi-temporal street-view imagery can capture fine-grained disaster damages that are often overlooked by top-down remote sensing. Wang et al.~\cite{Wang2025DisasterM3RemoteSensing} released DisasterM3, a large-scale remote sensing vision-language dataset that supports detailed reasoning about disaster events by aligning satellite imagery with textual descriptions.

Cross-view and temporal datasets provide the main empirical foundation for this shift. CVDisaster \cite{Li2025Crossviewgeolocalizationdisaster} pairs satellite and street-view observations for cross-view damage localization, while the Bi-Temporal StreetView Damage dataset \cite{Yang2025Hyperlocaldisasterdamagea} aligns pre- and post-disaster street views for change-based severity estimation. Recent generative work also explores satellite-to-street synthesis as a way to recover ground-level context when post-disaster street views are unavailable \cite{yang2026satellite}. Authoritative inspection and incident datasets add complementary supervision: the Los Angeles County Damage Inspection System provides georeferenced wildfire photographs with official severity ratings \cite{LosAngelesCountyFireDepartment2025DINS}, and the Incidents Dataset covers broader hazard categories from online post-disaster imagery \cite{Weber2020DetectingNaturalDisasters}. Social media and crowdsourced imagery further expand rapid disaster sensing, but they also introduce geographic bias and variable evidence quality \cite{Ma2025MultimodalMultilingualMultidimensionalb}.

\subsection{Agentic AI for Disaster Analysis}

Building on these advances in disaster image analysis, the rise of vision-language models (VLMs) and agentic AI has shifted disaster research from task-specific prediction toward systems that can perceive, reason, plan, and collaborate. Beyond recognizing damage, VLMs enable higher-level capabilities such as automatic disaster report generation \cite{Chen2024IntegrationLargeVision}, cross-modal question answering and contextual reasoning \cite{Sun2023UnleashingPotentialLarge}, social-media-based situation awareness \cite{Zhou2026RapidDisasterResponse, Zou2023SocialMediaEmergencyRescue}, disaster knowledge extraction and task planning \cite{Lei2025HarnessingLargeLanguagea}, and agent-based resilience frameworks \cite{Durante2024AgentAISurveying, Li2025TriEnvironmental}. These studies show that VLMs can process heterogeneous evidence from satellites, street views, and text streams, but most systems still target a single task rather than a complete assessment workflow.

Agentic systems offer a natural structure for such workflows, because disaster assessment can be decomposed into perception, restoration, recognition, reasoning, and reporting. Prior work has studied disaster perception and risk interpretation \cite{Cui2019Associationdisasterexperience, Ediz2024Disasterpreparednessperception, Ho2008Howdisastercharacteristics, Miceli2008Disasterpreparednessperception}, remote sensing and urban image restoration \cite{Chen2024IntegrationLargeVision, rasti2021image, Zuo20254KAgentAgenticAnya}, automated damage detection and classification \cite{Alisjahbana2024DeepDamageNettwostepdeeplearning, Gu2025Multiviewstreetview}, and large language model (LLM)-based geospatial disaster reasoning \cite{Han2025FineTuningLLMAssistedChinese, Rawat2024DisasterQABenchmarkAssessing, Wang2025Floodresiliencecities}. However, these capabilities are usually developed and evaluated in isolation rather than coordinated within a single autonomous system.

In summary, existing studies highlight a shift from hyperlocal sensing toward image-based understanding and, more recently, agent-based autonomy. Yet a gap remains between low-level fusion of remote sensing and street-view imagery and high-level autonomous reasoning. RAPID bridges this gap through a cross-view autonomous multi-agent framework for disaster diagnosis. By coordinating specialized agents for disaster perception, image restoration, damage recognition, and reasoning, the framework converts cross-view geospatial observations into structured and interpretable disaster intelligence.

\section{Methodology}

\subsection{RAPID: A Reproducible Multi-Agent Framework for Disaster Damage Intelligence}
As shown in Figure~\ref{fig:framework}, RAPID is a reproducible multi-agent system for cross-view disaster intelligence. The system integrates remote sensing imagery and street-view imagery into an end-to-end workflow for perception, image-quality control, structured damage analysis, reasoning, and report generation. Its four agents play distinct roles: the Disaster Perception Agent (DPA) identifies the imagery type and disaster context; the Image Restoration Agent (IRA) decides whether and how to restore degraded imagery; the Damage Recognition Agent (DRA) converts visual evidence into severity and object-level damage signals; and the Disaster Reasoning Agent (DReA) transforms these intermediate outputs into causal explanations and decision-support reports.

Let \(X=\{I_1,\ldots,I_n\}\) denote the visual evidence for one case. RAPID is implemented as a sequence of auditable state updates rather than a single black-box prediction:
\begin{equation}
\begin{aligned}
\mathbf{z}_{p} &= A_{\mathrm{DPA}}(X),\\
\tilde{X} &= A_{\mathrm{IRA}}(X,\mathbf{z}_{p}),\\
\mathbf{z}_{d} &= A_{\mathrm{DRA}}(\tilde{X},\mathbf{z}_{p}),\\
\mathcal{Y} &= A_{\mathrm{DReA}}(\mathbf{z}_{p},\mathbf{z}_{d},\tilde{X}).
\end{aligned}
\label{eq:pipeline}
\end{equation}
Here, \(\mathbf{z}_{p}\) stores perception and routing signals, \(\tilde{X}\) stores the audit-preserving image set passed downstream, \(\mathbf{z}_{d}\) stores structured damage outputs, and \(\mathcal{Y}\) is the final disaster report.

\subsubsection{Disaster Perception Agent}

DPA serves as the entry point of the pipeline. We use perception in the computer-vision sense of visual identification and scene parsing, not in the disaster-studies sense of human risk perception. DPA contains ModePerceiver, DisasterReasoner, and TaskPlanner. Together, they infer the input modality, disaster category, restoration need, and downstream tool route from a single multimodal prompt.

We define the perception state as
\begin{equation}
\mathbf{z}_{p}=(\hat{m},\hat{d},\hat{\mathbf{r}},\hat{c},\tau),
\qquad
\tau=\{\mathrm{DRA},\mathrm{DReA}\}\cup \{\mathrm{IRA}: \|\hat{\mathbf{r}}\|_1>0\},
\label{eq:dpa-state}
\end{equation}
where \(\hat{m}\in\mathcal{M}\) is the modality, \(\hat{d}\in\mathcal{D}\) is the disaster category, \(\hat{\mathbf{r}}\in\{0,1\}^{n}\) marks images requiring restoration, \(\hat{c}\) is confidence, and \(\tau\) is the planned agent route. This compact state replaces repeated free-form handoffs with a consistent routing interface.

\subsubsection{Image Restoration Agent}

IRA treats restoration as guarded quality control. For each image marked by DPA, it computes no-reference IQA features for contrast, sharpness, and naturalness. The composite score is
\begin{equation}
Q =
\frac{
w_c C_{\mathrm{norm}} + w_s S_{\mathrm{norm}} + w_n(1 - N_{\mathrm{norm}})
}{
w_c + w_s + w_n
}.
\label{eq:quality}
\end{equation}
Here, \(C_{\mathrm{norm}}\), \(S_{\mathrm{norm}}\), and \(N_{\mathrm{norm}}\) denote normalized contrast, sharpness, and naturalness-deviation terms. Higher \(Q\) indicates better perceptual quality for downstream interpretation. In the reported experiments we use \(w_c=0.4\), \(w_s=0.4\), and \(w_n=0.2\), prioritizing contrast and sharpness while still penalizing unnatural image statistics. These weights are fixed across datasets rather than tuned per disaster type.

The agent compares three branches: a deterministic baseline, a planner-guided deterministic tool chain, and an image-only Gemini enhancement. Let \(\mathcal{B}=\{\mathrm{baseline},\mathrm{planner},\mathrm{gemini}\}\). For branch \(b\), let \(\Delta_b=Q_b-Q_{\mathrm{original}}\). IRA selects the restored image only if the best branch improves quality by at least \(\delta\):
\begin{equation}
\tilde{I} =
\begin{cases}
I_{b^\star}, & \Delta_{b^\star}\ge\delta,\\
I, & \text{otherwise},
\end{cases}
\quad
b^\star=\arg\max_{b\in\mathcal{B}} \Delta_b.
\label{eq:restoration-selection}
\end{equation}
We set \(\delta=0.01\) for planner-guided restoration and require nonnegative improvement for the image-only branch. The original image remains available to DRA and human audit, so \(Q\) is used as a perceptual quality proxy rather than as proof of semantic faithfulness.

\subsubsection{Damage Recognition Agent}

DRA converts the audit-preserving image set into ordinal severity and object-level damage indicators. It supports cross-view RSI+SVI inputs, bi-temporal SVI pairs, and single post-disaster SVI. For \(K\) severity levels and \(J\) damage indicators, let \(\pi_{1:K}\in[0,1]^K\) denote severity confidences and \(\alpha_{1:J}\in[0,1]^J\) object confidences. The decision rule is
\begin{equation}
\hat{y}=\arg\max_{k\in\{0,\ldots,K-1\}}\pi_k,\qquad
\hat{o}_j=\mathbb{1}[\alpha_j\ge\theta_j],
\label{eq:dra-decision}
\end{equation}
where \(\theta_j\) is the threshold for the \(j\)-th damage element. The structured DRA state is \(\mathbf{z}_{d}=(\hat{y},\hat{\mathbf{o}},\pi_{1:K},\alpha_{1:J},\hat{\mathbf{b}},\hat{\mathbf{p}})\), where \(\hat{\mathbf{b}}\) and \(\hat{\mathbf{p}}\) are optional bounding boxes and polygon masks. This schema keeps cross-view, bi-temporal, and single-view cases comparable while preserving confidence information for reasoning.

We evaluate DRA with exact severity accuracy, object-level precision, recall, F1-score, and the Normalized Cross-Severity Error (NCSE), where the ``S'' denotes severity. NCSE complements categorical accuracy by penalizing large ordinal mistakes more strongly than adjacent-class errors.

\subsubsection{Disaster Reasoning Agent}

DReA maps structured recognition outputs to explanation and action. It conditions the report on perception state, damage state, image evidence, and a recovery-knowledge prompt \(\kappa\):
\begin{equation}
\mathcal{Y}_i = A_{\mathrm{DReA}}(\mathbf{z}_{p,i},\mathbf{z}_{d,i},\tilde{X}_i,\kappa).
\label{eq:drea-report}
\end{equation}
The report describes damage mechanisms, affected assets, secondary risks, and recovery suggestions such as debris clearance, vegetation restoration, structural safety inspection, and community risk mitigation.

Because \(\mathcal{Y}_i\) is open-ended text, we evaluate it with automated and human scores. Let \(\mathcal{R}\) denote four rubric dimensions: factual consistency, causal plausibility, completeness, and action feasibility. The automated judge is used as a first-pass rubric scorer rather than as ground truth, because LLM-based evaluation can favor familiar styles of reasoning. For human evaluation, three geography Ph.D. student reviewers independently score the sampled reports using the same 1--5 rubric after inspecting the associated image evidence and generated report. The final human score is the mean of their rubric scores across reviewers and dimensions:
\begin{equation}
\begin{aligned}
S_i^{\mathrm{auto}}
&= \frac{1}{|\mathcal{R}|}\sum_{r\in\mathcal{R}} s_{i,r}^{\mathrm{auto}},\\
S_i^{\mathrm{human}}
&= \frac{1}{3|\mathcal{R}|}\sum_{h=1}^{3}\sum_{r\in\mathcal{R}} s_{i,h,r}^{\mathrm{human}}.
\end{aligned}
\label{eq:reasoning-score}
\end{equation}
We summarize inter-reviewer reliability with four complementary statistics: mean pairwise absolute disagreement on the 1--5 scale, the percentage of paired ratings within one point, Krippendorff's ordinal alpha, and the average-measures intraclass correlation coefficient, ICC(2,k). These statistics evaluate whether the averaged human score is a stable summary of expert judgment rather than the opinion of a single reviewer.

\subsubsection{Execution Protocol and Audit Trail}

RAPID stores each case as an audit object that links inputs, intermediate outputs, predictions, and evaluation scores:
\begin{equation}
\mathcal{A}_i =
\{X_i,\tilde{X}_i,\mathbf{z}_{p,i},\mathbf{z}_{d,i},
\Delta_{i,1:|\mathcal{B}|},\mathcal{Y}_i,S_i^{\mathrm{auto}},S_i^{\mathrm{human}}\}.
\label{eq:audit-trail}
\end{equation}
Here, \(X_i\) denotes the original inputs, \(\tilde{X}_i\) denotes the selected restored inputs, \(\mathbf{z}_{p,i}\) and \(\mathbf{z}_{d,i}\) denote perception and damage outputs, \(\Delta_{i,1:|\mathcal{B}|}\) stores restoration-score changes across backbones, \(\mathcal{Y}_i\) denotes the generated report, and \(S_i^{\mathrm{auto}}\) and \(S_i^{\mathrm{human}}\) denote automatic and human evaluation scores. This record keeps visual evidence separate from model interpretation and allows errors to be traced to perception, restoration, damage recognition, or reasoning.

All model calls use zero-shot prompts with fixed label spaces, fixed output keys, and JSON-format constraints. DPA prompts enumerate disaster classes and modality labels. DRA prompts specify severity labels and object indicators. DReA prompts require generated reports to separate observed evidence, inferred risk, and recommended action. Invalid outputs are normalized to the closest valid schema category when possible. In this paper, zero-shot means that no target-dataset fine-tuning or few-shot examples are used in the prompt.

The pipeline separates operational routing from evaluation. IRA may improve image readability, but original images remain available for DRA and human review. DReA may suggest actions, but its output is evaluated as decision support rather than an autonomous emergency directive. Because commercial model endpoints can change, all runs should archive model names, endpoint versions, prompts, temperatures, raw outputs, call dates, code commit hashes, and serialized audit objects.

\subsection{Dataset and Study Area}
\begin{figure}[t]
  \centering
  \includegraphics[width=\columnwidth]{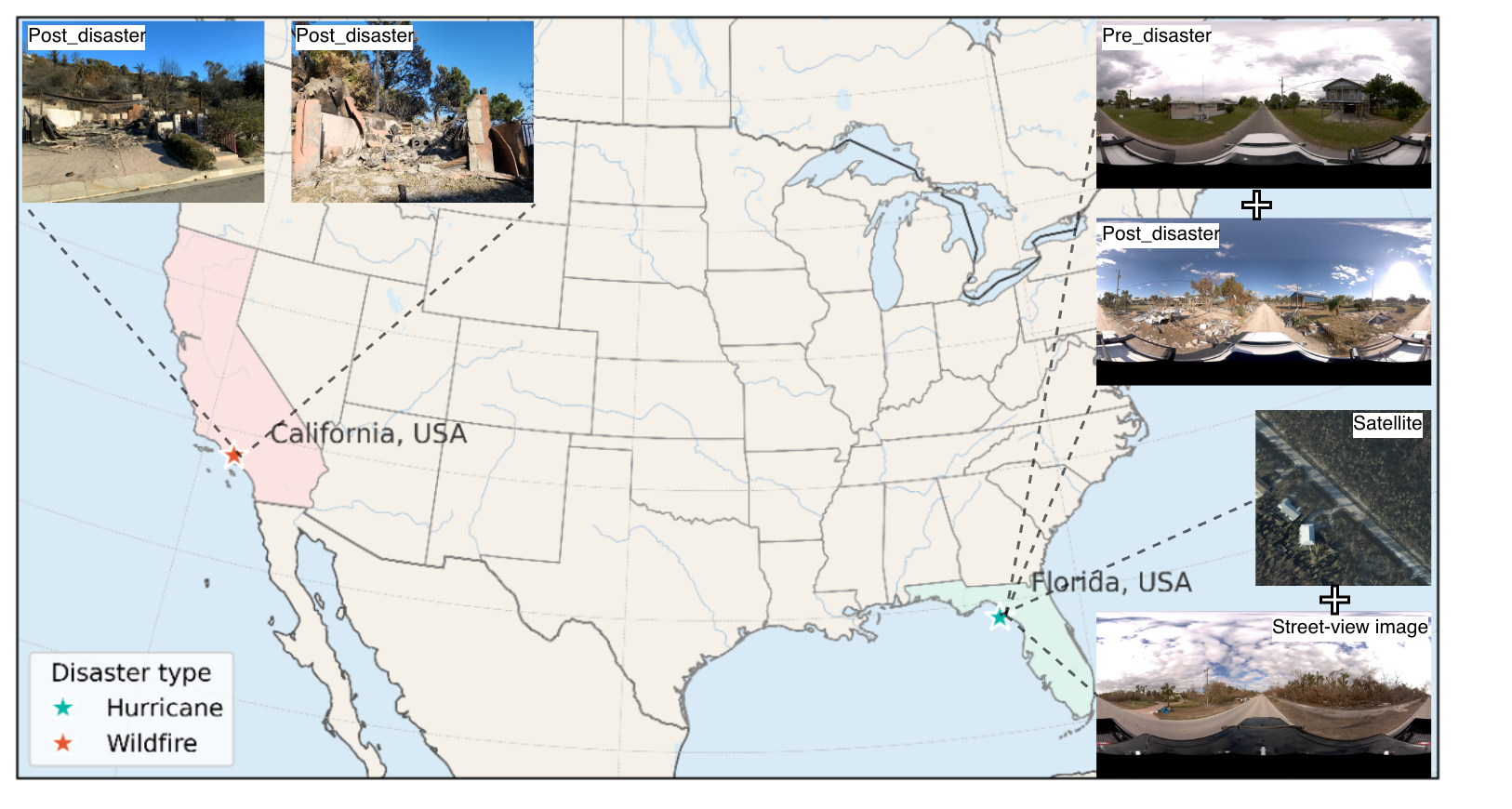}
  \caption{Geographic locations of disasters used in the evaluation.}
  \Description{Map and sample disaster images showing geographic coverage across California and Florida.}
  \label{fig:location}
\end{figure}

Based on the complementary characteristics of available disaster image datasets, we selected representative post-disaster datasets across three dimensions: cross-view, bi-temporal, and multi-hazard. The evaluation data comprise three core types of disaster images: (A) cross-view images from remote sensing and street-view imagery, (B) bi-temporal pre- and post-disaster street-view images, and (C) multi-hazard post-disaster street-view images. These three dataset types correspond to the three input modes supported by the Damage Recognition Agent. The georeferenced severity-labeled datasets mainly cover hurricanes and wildfires across California and Florida, while Dataset C is used for hazard-type perception rather than severity estimation. Figure~\ref{fig:location} shows the geographic distribution with coordinates and example images.

\begin{figure}[t]
  \centering
  \includegraphics[width=\columnwidth]{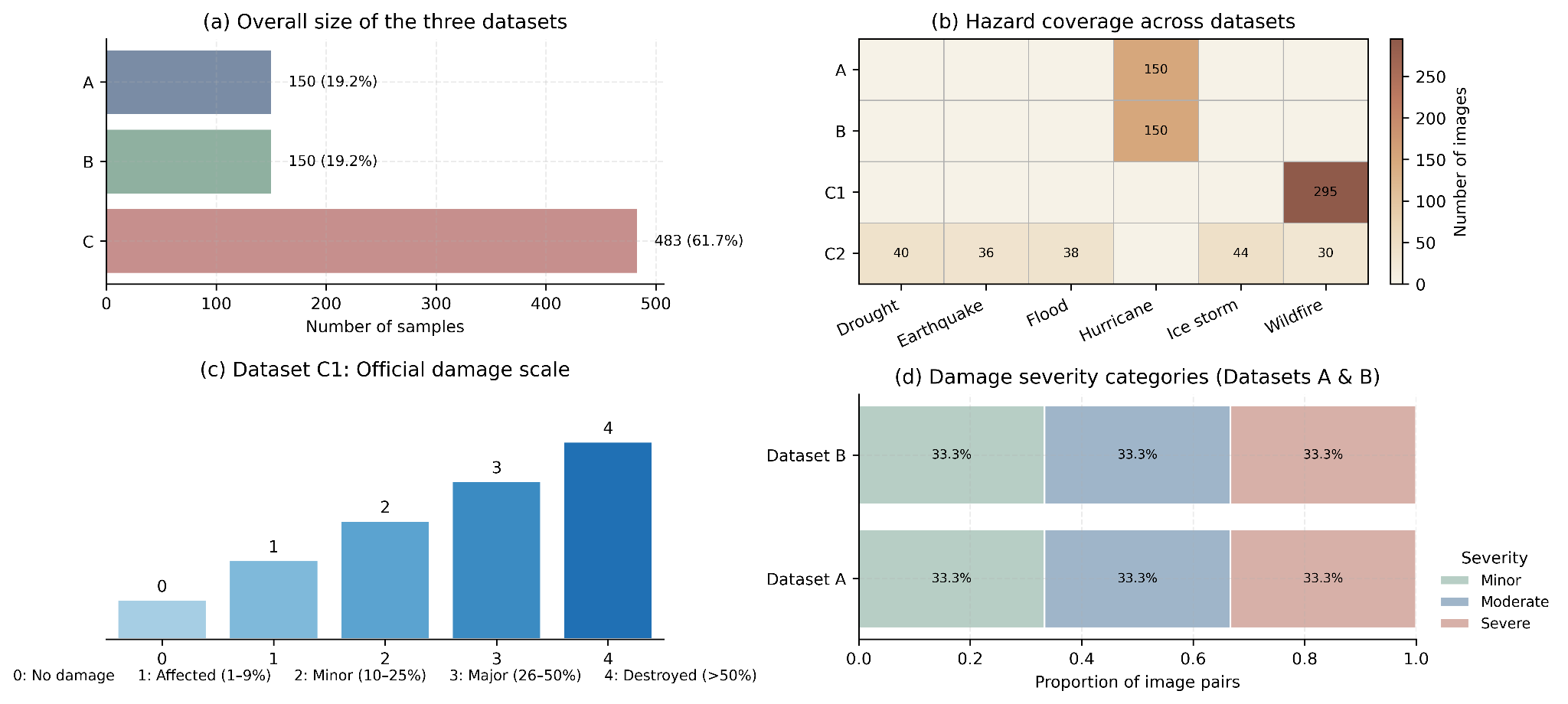}
  \caption{Key statistics of the dataset.}
  \Description{Charts comparing dataset size, disaster categories, and damage severity coverage.}
  \label{fig:stats}
\end{figure}

Datasets A and B each contain 150 image pairs. Dataset A is cross-view, containing synchronously acquired street-view and remote sensing images. Dataset B is bi-temporal and captures structural changes before and after a disaster event. Dataset C is larger, with post-disaster street-view images from multiple hazards and scenarios. These datasets provide complementary viewing structures, temporal structures, damage levels, and disaster scenarios for the multi-agent disaster diagnosis pipeline. Figure~\ref{fig:stats} summarizes the scale differences, disaster coverage, and official damage levels of the datasets. Expanding severity-labeled post-disaster street-view imagery is harder than collecting satellite scenes because ground-level images must be aligned across location, time, viewpoint, accessibility, and damage labels; road closures, safety constraints, platform coverage, privacy rules, licensing terms, and delayed revisits all reduce the number of public street-view observations that can be matched to official disaster severity.

Table~\ref{tab:datasets} summarizes the datasets used in this study. Dataset A consists of paired RSI and SVI for cross-view disaster analysis under a unified three-level severity standard. Dataset B includes bi-temporal street-view images for damage assessment from temporal change. Dataset C is divided into two subsets: C1 contains post-disaster street-view images from drought, earthquake, flood, hurricane, ice storm, and wildfire without explicit severity labels, while C2 focuses on wildfire scenarios using a five-level damage annotation scheme.

\begin{table*}[t]
\centering
\caption{Types, composition, and characteristics of the disaster imagery datasets used in this study.}
\label{tab:datasets}
\resizebox{\textwidth}{!}{%
\begin{tabular}{lllllll}
\toprule
\textbf{Dataset} & \textbf{Data Type} & \textbf{Images} & \textbf{Disaster} & \textbf{Severity} & \textbf{Source} & \textbf{Associated Agents} \\
\midrule
A & SVI+RSI pairs & 300 & Hurricane & 3 levels & CVDisaster \cite{Li2025Crossviewgeolocalizationdisaster} & DPA+IRA+DRA+DReA \\
B & Bi-temporal SVI & 300 & Hurricane & 3 levels & Bi-Temporal SVI \cite{Yang2025Hyperlocaldisasterdamagea} & DPA+IRA+DRA+DReA \\
C1 & Post-disaster SVI & 188 & Drought, earthquake, flood, hurricane, ice storm, wildfire & N/A & Incidents Dataset \cite{Weber2020DetectingNaturalDisasters} & DPA \\
C2 & Post-disaster SVI & 295 & Wildfire & 5 levels & LA DINS \cite{LosAngelesCountyFireDepartment2025DINS} & DPA+DRA \\
\bottomrule
\end{tabular}}
\end{table*}

\subsection{Scope, Assumptions, and Reproducibility Boundaries}

Our evaluation tests whether RAPID can organize heterogeneous visual evidence into auditable disaster information. It does not test whether RAPID can replace official field inspection. Three assumptions guide the study. First, each image provides only partial evidence: street-view imagery can reveal object-level damage, while satellite imagery provides broader spatial context. Second, severity labels are treated as ordered categories rather than continuous physical measurements, so DRA is evaluated with both exact accuracy and NCSE. Third, generated reports are treated as decision-support summaries whose claims must remain grounded in visible evidence and structured agent outputs.

The datasets support different evaluation tasks. Dataset A enables cross-view reasoning because remote sensing and street-view images describe the same event from complementary perspectives. Dataset B enables temporal reasoning by comparing pre- and post-disaster street-view images. Dataset C supports hazard recognition and wildfire severity analysis, but C1 lacks explicit severity labels. RAPID therefore activates only the agents that match the available evidence and labels, as shown in Table~\ref{tab:datasets}. This avoids forcing all datasets into a single task definition.

For reproducibility, each run stores four artifact types: visual evidence, agent outputs, evaluation metrics, and report text. Visual evidence includes the original images and selected restored images from Eq.~\ref{eq:restoration-selection}. Agent outputs include \(\mathbf{z}_{p}\), \(\mathbf{z}_{d}\), confidence scores, restoration deltas, and generated reports. Metrics are computed from stored outputs rather than interactive inspection. This design supports rerunning individual agents, auditing failure cases, and comparing model backbones without changing the dataset definition. Reproducibility is therefore procedural and audit-oriented, not bitwise deterministic, because proprietary APIs, model-version changes, and rate limits may affect exact outputs.

These boundaries shape how the results should be interpreted. A high DPA score indicates strong hazard-category recognition, but not necessarily correct severity estimation. A high restoration score indicates better perceptual quality under the selected IQA proxy, but it does not certify semantic faithfulness. A high DReA score indicates that a generated report aligns with the available evidence and rubric, but it remains a triage-oriented explanation rather than an official damage determination. RAPID is therefore best understood as an auditable early-stage assessment pipeline. The current datasets provide component-level evidence for the framework, not evidence of global disaster-response generalization.

\section{Experiments and Results}

\subsection{Evaluation Protocol}

The evaluation follows the modular structure in Eq.~\ref{eq:pipeline}. We evaluate DPA as a multi-class classification problem, IRA as a quality-control problem, DRA as severity and object-level recognition, and DReA as reasoning and report generation. This design avoids reducing the entire pipeline to one aggregate score, which would obscure which agent contributes to success or failure.

For DPA, the input is a post-disaster image or image pair, and the target is one of six disaster categories. Performance is reported with precision, recall, F1-score and overall accuracy. For IRA, the evaluation compares \(Q_{\mathrm{original}}\) with the branch-specific post-restoration scores. We emphasize the direction and stability of quality changes rather than treating restoration as ground-truth reconstruction, because a visually sharper image can still be semantically unsafe if disaster evidence is altered. For DRA, we use exact severity accuracy and NCSE for ordered labels, together with object-level precision, recall, F1-score, and binary accuracy for visible damage indicators. For DReA, we evaluate reports with both automated LLM scoring and human review using the same rubric dimensions.

Across all tasks, the models are used in a zero-shot setting. No model is fine-tuned on the target labels, and the same label definitions are reused across comparable datasets. This setting reflects early disaster-response conditions, where labeled samples from the current event are limited or unavailable. It also makes model differences easier to interpret: performance changes are more directly tied to the input modality, temporal structure, and reasoning prompt rather than to dataset-specific retraining. The reported model-backbone comparisons are intended as zero-shot model and evidence-structure checks, not as a full supervised benchmark study. Representative supervised change-detection and building-damage systems usually require event-specific labels, sensor geometries, and annotation taxonomies; applying them to a new region would require a separate adaptation protocol. RAPID is therefore evaluated as an early-response workflow for scarce-label settings, while supervised baselines remain most appropriate for future in-domain or transfer-learning comparisons. A component-removal ablation also requires care because removing DPA changes routing, removing IRA changes image-quality control only for degraded cases, and removing DReA changes the output from report generation to structured prediction. The present evaluation, therefore, reports module-level behavior and audit states rather than collapsing all agents into one task definition.

Because several datasets are small, we interpret point estimates with caution. Approximate Wilson 95\% confidence intervals are about 0.87--0.95 for DPA accuracy on C1, 0.55--0.70 for the best Dataset A severity accuracy, 0.51--0.67 for the best Dataset B severity accuracy, and 0.53--0.65 for the best Dataset C2 severity accuracy. These intervals suggest that large performance differences are meaningful, but small gaps between models should not be overinterpreted.

\subsection{Evaluation of the Disaster Perception Agent on Multi-Disaster Classification}

We evaluate DPA on a multi-disaster classification task, where each instance is assigned to one of six hazard categories: drought, earthquake, flood, hurricane, ice storm, and wildfire. We compare three LLM backbones and report per-class precision, recall, F1-score, and overall accuracy in Table~\ref{tab:perception}. All models show strong classification ability and consistently achieve high precision across most hazard categories, usually surpassing 0.90. Differences mainly stem from recall, which indicates sensitivity to more difficult or uncertain cases. Earthquake classification remains the most challenging category, likely because of greater visual variability and weaker distinguishing features. 

Among the evaluated models, GPT-5-mini achieves the best overall performance, with an accuracy of 0.92, surpassing ChatGPT-5.1 at 0.88 and Gemini-2.5-flash at 0.86. It also shows the most balanced precision and recall, earning F1-scores of at least 0.94 for drought, flood, hurricane, and wildfire, while maintaining the highest recall for earthquake and ice storm categories. The gap between overall accuracy and earthquake recall also shows why per-class metrics are needed for imbalanced multi-disaster perception.

\begin{table}[t]
\centering
\caption{Performance comparison of the Disaster Perception Agent in multi-disaster type classification.}
\label{tab:perception}
\makebox[\columnwidth][c]{%
\resizebox{1.05\columnwidth}{!}{%
\begin{tabular}{llcccccc}
\toprule
\textbf{Metric} & \textbf{Model} & \textbf{Drought} & \textbf{Earthquake} & \textbf{Flood} & \textbf{Hurricane} & \textbf{Ice storm} & \textbf{Wildfire} \\
\midrule
\multirow{3}{*}{Precision \(\uparrow\)}
& GPT-5-mini & 0.95 & 0.91 & 0.90 & \textbf{0.98} & \textbf{1.00} & \textbf{0.99} \\
& ChatGPT-5.1 & \textbf{0.97} & \textbf{0.95} & \textbf{0.92} & 0.97 & 0.97 & \textbf{0.99} \\
& Gemini-2.5-flash & 0.95 & \textbf{0.95} & 0.90 & 0.91 & \textbf{1.00} & \textbf{0.99} \\
\midrule
\multirow{3}{*}{Recall \(\uparrow\)}
& GPT-5-mini & \textbf{0.93} & \textbf{0.58} & \textbf{0.97} & 0.99 & \textbf{0.86} & 0.98 \\
& ChatGPT-5.1 & 0.85 & 0.56 & 0.92 & \textbf{1.00} & 0.70 & 0.97 \\
& Gemini-2.5-flash & \textbf{0.93} & 0.50 & 0.92 & \textbf{1.00} & 0.50 & \textbf{0.99} \\
\midrule
\multirow{3}{*}{F1-score \(\uparrow\)}
& GPT-5-mini & \textbf{0.94} & \textbf{0.71} & \textbf{0.94} & \textbf{0.99} & \textbf{0.93} & 0.98 \\
& ChatGPT-5.1 & 0.91 & 0.70 & 0.92 & \textbf{0.99} & 0.82 & 0.98 \\
& Gemini-2.5-flash & \textbf{0.94} & 0.65 & 0.91 & 0.95 & 0.67 & \textbf{0.99} \\
\midrule
\multirow{3}{*}{Overall accuracy \(\uparrow\)}
& GPT-5-mini & \multicolumn{6}{c}{\textbf{0.92}} \\
& ChatGPT-5.1 & \multicolumn{6}{c}{0.88} \\
& Gemini-2.5-flash & \multicolumn{6}{c}{0.86} \\
\bottomrule
\end{tabular}}%
}
\end{table}

\subsection{Evaluation of the Image Restoration Agent for Visual Quality Enhancement}

The Image Restoration Agent generally enhances the visual quality of post-disaster images, although the extent and ranking of improvements vary across image types and disaster scenarios, as shown in Table~\ref{tab:restoration}. Here, \(Q_{\mathrm{original}}\) represents the quality score before restoration, while \(Q_{\mathrm{baseline}}\), \(Q_{\mathrm{planner}}\), and \(Q_{\mathrm{gemini}}\) represent the post-restoration scores for the baseline, planner-guided, and Gemini image-only branches, respectively. The number of restored images is reported explicitly because several subsets contain only a few degraded inputs; those rows should be read as diagnostic evidence rather than statistically powered restoration benchmarks.

For hurricane satellite imagery, the baseline and planner branches deliver the largest improvements, raising the score from 0.62 to 0.73 and 0.71, respectively, while \(Q_{\mathrm{gemini}}\) reaches 0.69. This indicates that structured or guided restoration methods are more effective for large-scale, top-down imagery, where maintaining global consistency and texture is essential. For hurricane street-view imagery, the Gemini branch attains the highest score, increasing quality from 0.75 to 0.79. This suggests that generative enhancement can be effective for certain ground-level scenes; however, as shown below, this advantage does not hold uniformly across hazards.

Across the remaining post-disaster street-view subsets, \(Q_{\mathrm{planner}}\) attains the highest or nearly highest values, showing the most stable performance under diverse conditions, with \(Q_{\mathrm{baseline}}\)  typically close behind. In several cases (e.g., wildfire, ice storm, and drought), however, \(Q_{\mathrm{gemini}}\) underperforms both \(Q_{\mathrm{baseline}}\) and \(Q_{\mathrm{planner}}\), and can even drop below \(Q_{\mathrm{original}}\). This indicates that generative enhancement, despite its strength on some street-view scenes, exhibits higher variance and may introduce artifacts or distort disaster-relevant features in visually complex or degraded scenarios.

\begin{table*}[t]
\centering
\caption{Quantitative evaluation of image restoration across remote sensing and street-view disaster imagery.}
\label{tab:restoration}
\resizebox{\textwidth}{!}{%
\begin{tabular}{llcccccc}
\toprule
\textbf{Category} & \textbf{Disaster type} & \textbf{Image type} & \textbf{Restored images} & \(\mathbf{Q_{\mathrm{original}}}\) & \(\mathbf{Q_{\mathrm{baseline}}}\) & \(\mathbf{Q_{\mathrm{planner}}}\) & \(\mathbf{Q_{\mathrm{gemini}}}\) \\
\midrule
A: Post-disaster SVI+RSI pairs & Hurricane & Satellite & 141 & 0.62 & \textbf{0.73} & 0.71 & 0.69 \\
A: Post-disaster SVI+RSI pairs & Hurricane & SVI & 22 & 0.75 & 0.78 & 0.76 & \textbf{0.79} \\
B: Pre-/post-disaster SVI pairs & Hurricane & SVI & 4 & 0.76 & 0.78 & 0.78 & \textbf{0.79} \\
C2: Post-disaster SVI & Wildfire & SVI & 14 & 0.61 & \textbf{0.67} & \textbf{0.67} & 0.57 \\
C1: Post-disaster SVI & Hurricane & SVI & 2 & 0.51 & 0.55 & \textbf{0.62} & 0.60 \\
C1: Post-disaster SVI & Ice storm & SVI & 4 & 0.73 & 0.74 & \textbf{0.75} & 0.67 \\
C1: Post-disaster SVI & Flood & SVI & 2 & 0.68 & 0.73 & \textbf{0.77} & 0.71 \\
C1: Post-disaster SVI & Earthquake & SVI & 2 & 0.71 & 0.79 & \textbf{0.80} & 0.77 \\
C1: Post-disaster SVI & Drought & SVI & 2 & 0.66 & 0.75 & \textbf{0.78} & 0.61 \\
C1: Post-disaster SVI & Wildfire & SVI & 3 & 0.70 & 0.77 & \textbf{0.78} & \textbf{0.78} \\
\bottomrule
\end{tabular}}
\end{table*}

\begin{figure}[t]
  \centering
  \includegraphics[width=\columnwidth]{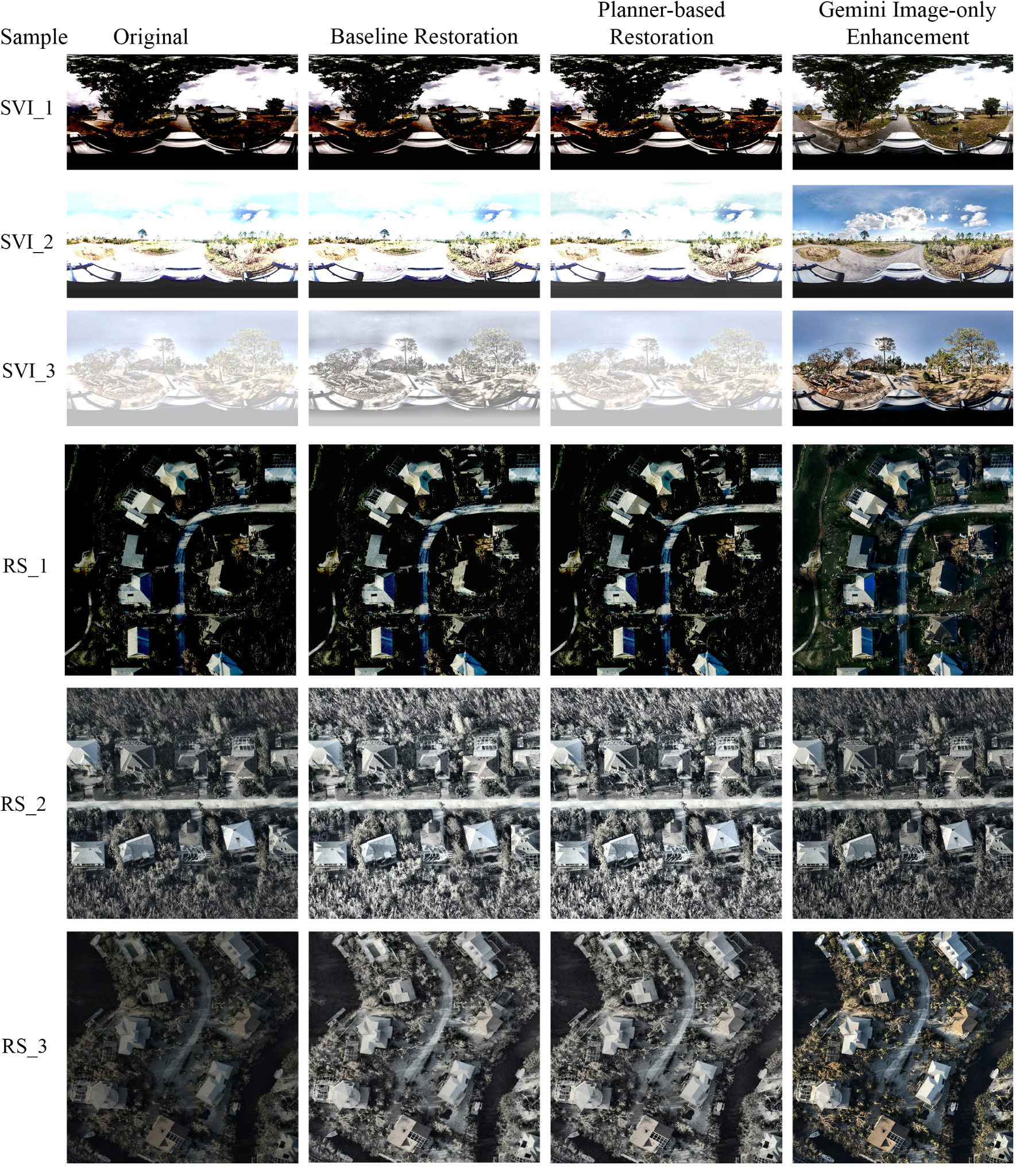}
  \caption{Visual comparison of restoration outputs produced by the Image Restoration Agent across baseline enhancement, planner-based restoration, and Gemini image-only optimization for SVI and RSI samples.}
  \Description{Side-by-side restored disaster images comparing baseline, planner-based, and Gemini image-only restoration outputs.}
  \label{fig:restoration}
\end{figure}

Figure~\ref{fig:restoration} compares restoration results across three branches for street-view and remote sensing samples. Visual observations align with Table~\ref{tab:restoration}: Gemini-based enhancement can improve exposure and texture in some street-view scenes, while baseline and planner-based methods better reduce haze, enhance contrast, and clarify structure in remote sensing images. The findings favor an adaptive, multi-branch restoration strategy over a single approach. They do not, however, establish semantic restoration faithfulness. The three-reviewer factual-consistency evaluation of DReA reports checks whether final reports remain aligned with image evidence, but it is not a restoration-specific semantic equivalence test. A practical deployment should compare DRA outputs on original and restored images, flag inconsistent damage evidence, and route high-discrepancy cases to human review.

\subsection{ Evaluation of the Damage Recognition Agent for Severity and Object-Level Damage Recognition}

We evaluate DRA on multi-disaster severity prediction using overall accuracy and NCSE. Let \(y_i, \hat{y}_i \in \{0,\ldots,K-1\}\) denote the ground-truth and predicted damage class for sample \(i\), where larger values indicate more severe damage and \(K\) is the number of ordered categories. NCSE is defined as
\begin{equation}
  \mathrm{NCSE} = \frac{1}{N}\sum_{i=1}^{N}\frac{\left|y_i - \hat{y}_i\right|}{K-1}.
\end{equation}
Here, \(N\) is the total number of samples. NCSE ranges from 0, indicating perfect agreement, to 1, indicating that all predictions fall in the most distant class. It penalizes large severity gaps more strongly than confusions between adjacent categories.

Table~\ref{tab:severity} shows DRA performance across three severity-labeled datasets. Accuracy is computed as exact agreement between predicted and reference severity labels, while NCSE captures how far an incorrect prediction falls from the correct ordinal class. For hurricane post-disaster SVI and RSI pairs, Gemini-3-Pro achieves the best performance, with an accuracy of 0.627 and the lowest NCSE of 0.190. This result is above a three-class random baseline, but it still leaves a substantial fraction of exact severity labels incorrect and should be interpreted as triage-level severity estimation rather than operational certification. For the bi-temporal SVI dataset, ChatGPT-5.1 attains the highest accuracy of 0.591 and the lowest NCSE of 0.218, suggesting the advantage of temporal change cues. In the wildfire C2 dataset with post-disaster SVI, Gemini-2.5-Pro slightly outperforms other models, achieving an accuracy of 0.590 and the lowest NCSE of 0.162.

\begin{table}[t]
\centering
\caption{Performance comparison of the Damage Recognition Agent in multi-disaster severity levels.}
\label{tab:severity}
\resizebox{\columnwidth}{!}{%
\begin{tabular}{llccc}
\toprule
\textbf{Metric} & \textbf{Model} & \textbf{Dataset A} & \textbf{Dataset B} & \textbf{Dataset C2} \\
\midrule
\multirow{5}{*}{Accuracy \(\uparrow\)}
& GPT-5-mini & 0.387 & 0.503 & 0.573 \\
& ChatGPT-5.1 & 0.573 & \textbf{0.591} & 0.570 \\
& Gemini-2.5-flash & 0.360 & 0.470 & 0.559 \\
& Gemini-2.5-Pro & 0.380 & 0.447 & \textbf{0.590} \\
& Gemini-3-Pro & \textbf{0.627} & 0.493 & 0.442 \\
\midrule
\multirow{5}{*}{NCSE \(\downarrow\)}
& GPT-5-mini & 0.307 & 0.248 & 0.165 \\
& ChatGPT-5.1 & 0.213 & \textbf{0.218} & 0.173 \\
& Gemini-2.5-flash & 0.363 & 0.299 & 0.179 \\
& Gemini-2.5-Pro & 0.373 & 0.327 & \textbf{0.162} \\
& Gemini-3-Pro & \textbf{0.190} & 0.291 & 0.210 \\
\bottomrule
\end{tabular}}
\end{table}

\begin{figure}[t]
  \centering
  \includegraphics[width=\columnwidth]{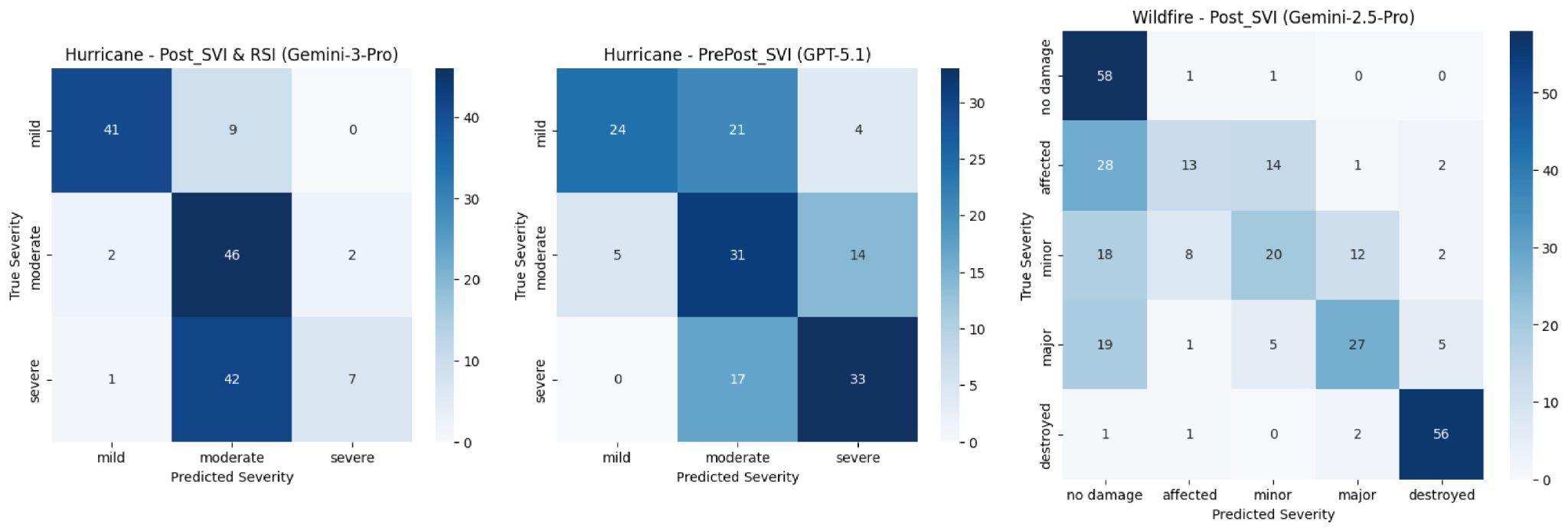}
  \caption{Confusion matrices of the best-performing model across three datasets.}
  \Description{Three confusion-matrix panels showing severity-level prediction errors across the evaluated datasets.}
  \label{fig:confusion}
\end{figure}

The confusion matrices in Figure~\ref{fig:confusion} show a consistent trend across datasets: errors mainly occur between neighboring severity levels rather than at the extremes. This suggests that models recognize a continuous damage spectrum but have difficulty with precise boundary separation in intermediate classes. Extreme categories, such as no damage and destroyed, are easier to distinguish.

Table~\ref{tab:object} compares disaster damage recognition performance under two evaluation settings. Dataset A focuses on cross-view post-disaster analysis, where the model must jointly infer damage conditions from remote sensing imagery and street-view images. Dataset B emphasizes bi-temporal street-view analysis, where the model compares pre-disaster and post-disaster observations to assess disaster impacts from visual changes.

In this table, the first four numerical columns report aggregate multilabel performance, while the final five columns report element-wise binary accuracy for the corresponding damage indicators. These per-indicator accuracies should be read together with F1-score, recall, and precision because rare indicators can yield high binary accuracy even when positive detections remain difficult.

For Dataset A, Gemini-3-Pro demonstrates the strongest overall performance with the highest F1-score of 0.77, precision of 0.79, and overall accuracy of 0.92. It also performs best on several key damage elements, including damaged buildings at 0.94, downed power lines at 1.00, and fallen trees at 0.92. By contrast, Gemini-2.5-Pro and Gemini-2.5-flash achieve relatively high recall but lower precision, suggesting that they broadly identify potential damage elements but also produce more false positives. ChatGPT-5.1 is more conservative and achieves higher overall correctness than both Gemini-2.5 variants, but still falls short of Gemini-3-Pro.

The pattern shifts in Dataset B, where ChatGPT-5.1 becomes the top-performing model across all major metrics. It achieves an F1-score of 0.55, recall of 0.53, precision of 0.57, and overall accuracy of 0.95. This advantage is important because the bi-temporal setting requires the model to identify and interpret changes between pre-disaster and post-disaster street-view images, distinguishing true disaster-related changes from variations in lighting, viewpoint, and environmental conditions.

\begin{table*}[t]
\centering
\caption{Quantitative performance of different models on disaster damage recognition across cross-view hurricane imagery (Dataset A) and bi-temporal street-view imagery (Dataset B).}
\label{tab:object}
\resizebox{\textwidth}{!}{%
\begin{tabular}{llccccccccc}
\toprule
\textbf{Category} & \textbf{Model} & \textbf{F1-score} & \textbf{Recall} & \textbf{Precision} & \textbf{Accuracy} & \makecell{\textbf{Damaged}\\\textbf{Building}} & \textbf{Debris} & \makecell{\textbf{Downed}\\\textbf{Power Lines}} & \makecell{\textbf{Fallen}\\\textbf{Tree}} & \makecell{\textbf{Flooded}\\\textbf{Area}} \\
\midrule
\multirow{5}{*}{\makecell[l]{Dataset A:\\SVI+RSI pairs}}
& Gemini-3-Pro & \textbf{0.77} & 0.78 & \textbf{0.79} & \textbf{0.92} & \textbf{0.94} & \textbf{0.89} & \textbf{1.00} & \textbf{0.92} & 0.88 \\
& Gemini-2.5-Pro & 0.57 & \textbf{0.91} & 0.50 & 0.71 & 0.57 & 0.77 & 0.67 & 0.63 & \textbf{0.95} \\
& Gemini-2.5-flash & 0.55 & 0.81 & 0.53 & 0.74 & 0.53 & 0.76 & 0.83 & 0.67 & 0.94 \\
& ChatGPT-5.1 & 0.54 & 0.58 & 0.64 & 0.84 & 0.74 & 0.84 & 0.97 & 0.74 & 0.91 \\
& GPT-5-mini & 0.49 & 0.57 & 0.46 & 0.79 & 0.75 & 0.81 & 0.92 & 0.60 & 0.90 \\
\midrule
\multirow{5}{*}{\makecell[l]{Dataset B:\\pre-/post-SVI pairs}}
& Gemini-3-Pro & 0.39 & 0.35 & 0.51 & 0.84 & 0.74 & 0.80 & 0.99 & 0.72 & \textbf{0.99} \\
& Gemini-2.5-Pro & 0.42 & 0.46 & 0.39 & 0.78 & 0.72 & 0.79 & 0.77 & 0.67 & \textbf{0.99} \\
& Gemini-2.5-flash & 0.43 & 0.43 & 0.43 & 0.83 & 0.78 & 0.78 & 0.93 & 0.68 & 0.98 \\
& ChatGPT-5.1 & \textbf{0.55} & \textbf{0.53} & \textbf{0.57} & \textbf{0.95} & \textbf{0.97} & \textbf{0.92} & \textbf{1.00} & \textbf{0.91} & \textbf{0.99} \\
& GPT-5-mini & 0.46 & 0.48 & 0.45 & 0.86 & 0.78 & 0.82 & 0.97 & 0.74 & \textbf{0.99} \\
\bottomrule
\end{tabular}}
\end{table*}

Another key factor influencing both datasets is class imbalance. Certain damage elements, such as flooded areas and downed power lines, occur less frequently, while debris and fallen trees are more common. This imbalance can make minority categories harder to model reliably. The results nevertheless show a clear overall pattern: different models have specific strengths depending on the task setting, and no single backbone is consistently optimal across all scenarios.

\begin{figure}[t]
  \centering
  \includegraphics[width=\columnwidth]{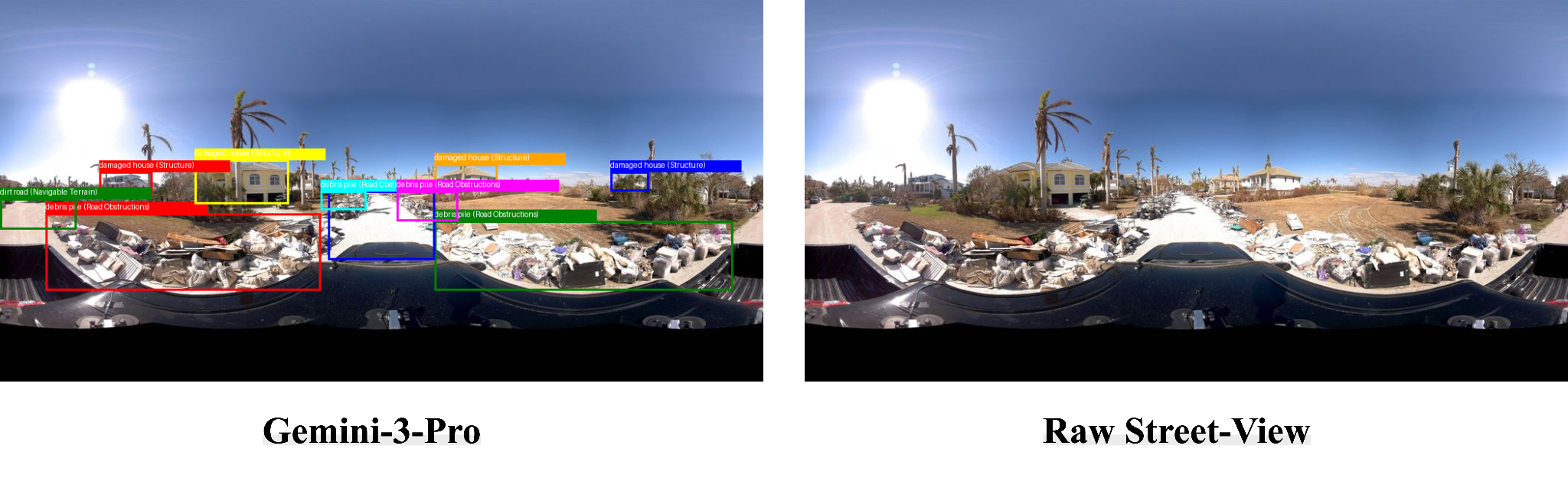}
  \caption{Object detection results of Gemini-3-Pro on a post-disaster street-view image, with the raw input shown for reference.}
  \Description{Object-detection visualization showing disaster-related objects such as debris, fallen trees, and damaged buildings in a street-view scene.}
  \label{fig:object_detection}
\end{figure}

Figure~\ref{fig:object_detection} shows object detection results from Gemini-3-Pro on a typical post-disaster street-view scene. The model identifies key disaster-related elements, such as damaged buildings, debris, fallen trees, and critical infrastructure, using a consistent detection standard. The predicted bounding boxes are spatially aligned and match the scene structure, suggesting plausible localization and semantic coherence.

\subsection{Evaluation of the Disaster Reasoning Agent for Evidence-Grounded Reasoning and Report Generation}

\begin{figure*}[t]
  \centering
  \includegraphics[width=0.85\textwidth]{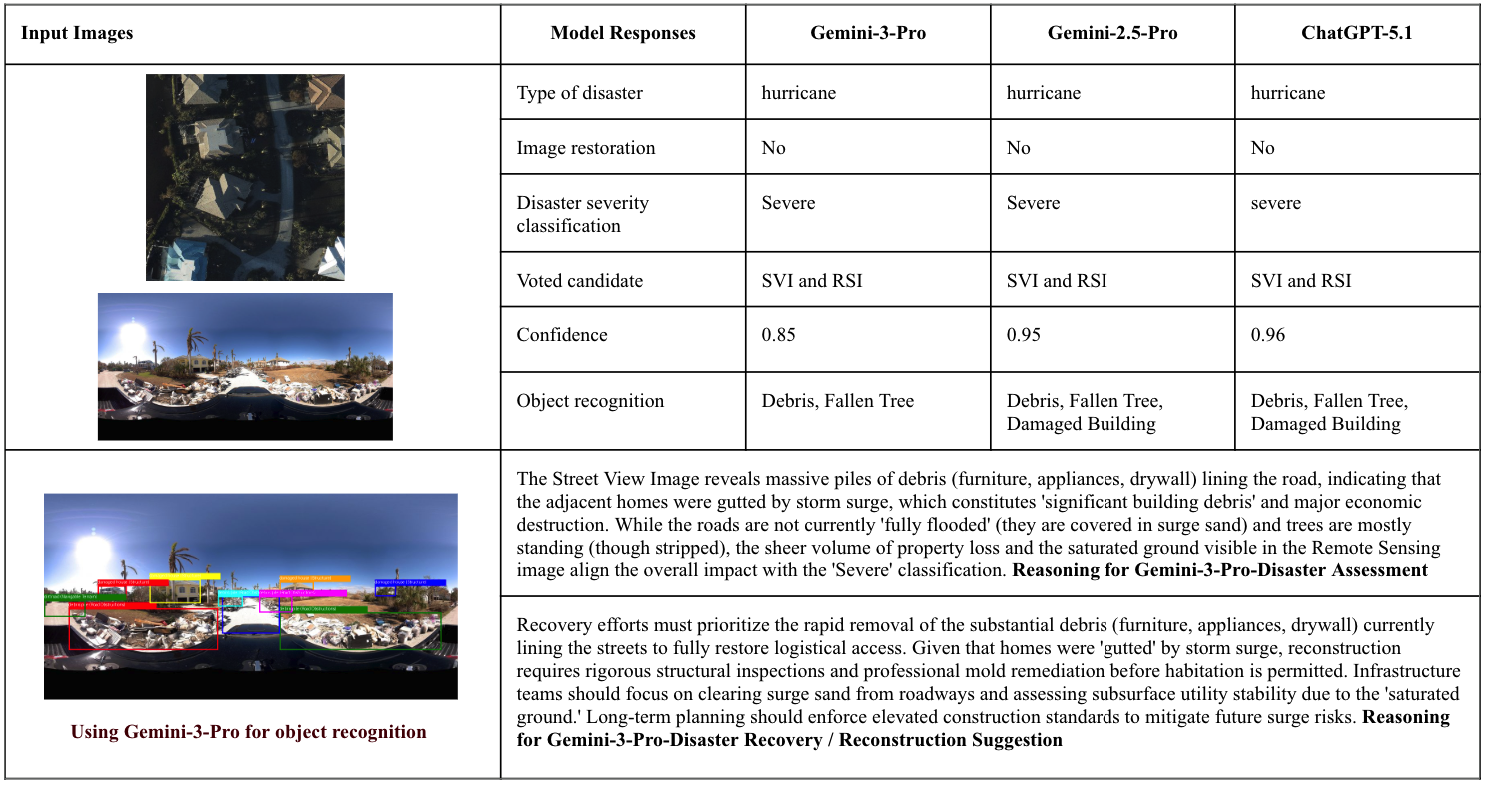}
  \caption{Comparative cross-view disaster assessment and reasoning results across Gemini-3-Pro, Gemini-2.5-Pro, and ChatGPT-5.1.}
  \Description{Comparative reasoning report showing disaster type, damage level, key object detection, confidence, and recommended actions across three vision-language models.}
  \label{fig:reasoning_results}
\end{figure*}

Figure~\ref{fig:reasoning_results} illustrates an example output from DReA. Using combined remote sensing and street-view images as inputs, the figure compares results from Gemini-3-Pro, Gemini-2.5-Pro, and ChatGPT-5.1 in terms of disaster type identification, damage level assessment, key object detection, model confidence, and disaster reasoning. Based on the model output, DReA combines information from street-view images, which show extensive debris, fallen trees, and structural damage, with remote sensing images, which indicate saturated ground conditions. It explains the causes and extent of the disaster from a causal-chain perspective and suggests short-term recovery actions such as clearing road debris, conducting structural safety assessments, performing mold remediation, and checking infrastructure stability.

\begin{figure}[t]
  \centering
  \includegraphics[width=\columnwidth]{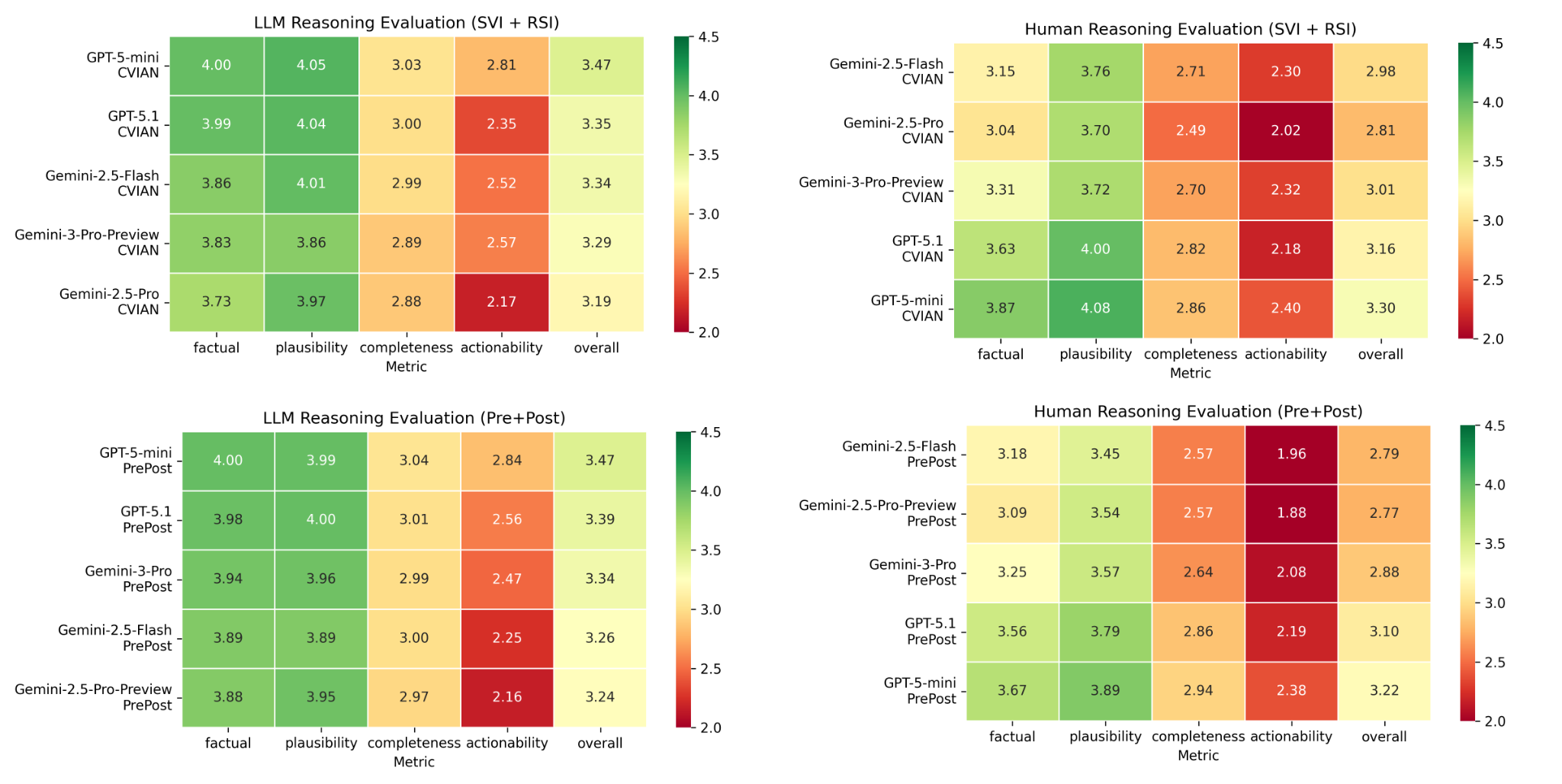}
  \caption{LLM and human evaluation of disaster reasoning.}
  \Description{Bar charts comparing LLM-based and human evaluation scores for factual consistency, causal plausibility, completeness, action feasibility, and overall reasoning quality.}
  \label{fig:reasoning_eval}
\end{figure}

Figure~\ref{fig:reasoning_eval} shows the results of reasoning-quality assessment for disaster reports generated by several vision-language models on two dataset types, comparing both automatic LLM evaluation and manual review by three geography Ph.D. student reviewers. The four dimensions are factual consistency, causal plausibility, completeness, and action feasibility. Scores range from 1 to 5, where higher values indicate stronger agreement with image evidence and recovery usefulness. Automated scores are treated as rubric-guided approximations; they may contain self-preference or style-preference bias and therefore are interpreted alongside human scores rather than alone.

Overall, scores for factual consistency and causal plausibility are generally high, with most models scoring close to or above 4.0 in these dimensions. This suggests that current models can maintain basic factual accuracy and causal relationships in disaster scenario reasoning. Scores for completeness and action feasibility are lower. In particular, action-feasibility scores typically range from 2.0 to 2.8, indicating that although models can describe disaster situations and offer some explanations, they still have shortcomings in proposing specific recovery measures, decision-making suggestions, or feasible solutions. Across datasets, combining street-view and remote sensing data (the cross-view configuration) generally outperforms the bi-temporal pre-/post-disaster configuration in terms of completeness and overall score. Human assessment results largely align with LLM assessment trends: factual consistency and causal plausibility scores are high, while action feasibility remains the lowest. Inter-reviewer reliability is substantial: the mean pairwise absolute disagreement is 0.34 points on the 1--5 scale, 96.1\% of paired ratings differ by no more than one point, Krippendorff's ordinal alpha is 0.76, and ICC(2,k) is 0.84 for the averaged human score. Agreement is strongest for factual consistency (\(\alpha=0.81\)) and causal plausibility (\(\alpha=0.79\)), remains acceptable for completeness (\(\alpha=0.73\)), and is lowest for action feasibility (\(\alpha=0.68\)), which is expected because feasible recovery recommendations depend more strongly on local operational context. This pattern is important for deployment: RAPID can summarize evidence, but response prioritization, infrastructure closure, or resource allocation still require expert review and local operational constraints.

\section{Discussion}
This study shows that RAPID can support an end-to-end cross-view disaster assessment workflow without task-specific fine-tuning. Instead of treating disaster assessment as a single prediction task, RAPID decomposes it into perception, image restoration, structured damage recognition, and high-level reasoning. This design makes intermediate outputs inspectable and allows different agents and model backbones to contribute where they are most effective.

The results suggest that RAPID's strength comes from coordination across components rather than from any single model, with each agent showing complementary strengths and task-dependent limitations.

The Disaster Perception Agent is the most reliable stage, exceeding 0.90 precision on most hazards and reaching 0.92 overall accuracy, but its recall drops to 0.50--0.58 for visually ambiguous categories such as earthquakes. The Image Restoration Agent shows that no single strategy dominates: controlled and planner-based methods are the most stable for remote sensing and most street-view subsets, while generative enhancement helps certain scenes (hurricane street-view, 0.75 to 0.79) yet can degrade quality below the original in wildfire, ice-storm, and drought cases. The Damage Recognition Agent reveals complementary backbones—Gemini-3-Pro leads cross-view recognition (0.627 severity accuracy, 0.77 object F1) and GPT-5.1 leads the bi-temporal setting (0.591, 0.55)—but exact severity accuracy remains modest (about 0.59--0.63), and errors cluster on adjacent severity levels rather than extremes. At the reasoning level, the Disaster Reasoning Agent scores highly on factual consistency and causal plausibility (close to or above 4.0 under both automatic and human evaluation), but completeness and especially action feasibility lag behind (2.0--2.8), with action feasibility also showing the lowest inter-reviewer agreement (\(\alpha=0.68\)). This shows that RAPID is dependable for evidence summarization and causal explanation, yet operational decisions still require expert judgment and local context.

RAPID still faces several challenges. First, its reliance on proprietary vision-language models creates constraints in cost, latency, and reproducibility. Second, the current framework does not yet include systematic evaluation of bias, fairness, or privacy risks. Third, severity prediction errors and inconsistencies in ground-truth labels indicate that uncertainty exists in both model outputs and evaluation data. These issues point to the need for a stronger human-AI collaborative workflow in which model predictions, uncertainty estimates, and expert judgment are considered together.

Future work should reduce dependence on closed models, extend RAPID to real-time data streams and broader disaster types, and integrate uncertainty quantification, human-in-the-loop validation, and responsible reasoning mechanisms. These extensions would move RAPID from an auditable descriptive analysis system toward a more reliable decision-support framework for disaster response.

\section{Conclusion}
We introduced RAPID, a reproducible multi-agent framework for zero-shot disaster damage assessment across cross-view geospatial data. By decomposing assessment into four coordinated agents—perception, image restoration, damage recognition, and reasoning—RAPID forms an end-to-end, interpretable workflow over remote sensing, street-view, and bi-temporal imagery.

Across multiple datasets and disaster types, RAPID performs hazard identification, severity estimation, and object-level damage recognition without task-specific fine-tuning, while exposing inspectable intermediate states that single-model pipelines hide. It thus reduces reliance on large-scale retraining and manual annotation, and shifts disaster assessment from pattern recognition toward auditable, evidence-grounded reasoning. RAPID does not replace official inspection or expert judgment; it provides an early-stage, decision-support workflow for time-sensitive response.

Its main limitations are a dependence on proprietary models and reasoning that is stronger in factual description than in concrete operational recommendation; uncertainty quantification, human-in-the-loop validation, and responsible evaluation of bias, fairness, and privacy remain open. Addressing these—through controllable open-source models, streaming data, broader hazard coverage, and tighter expert coupling—would move RAPID from an auditable descriptive system toward a reliable decision-support framework for the early stages of disaster response.

\begin{acks}
The authors used AI-assisted writing tools, including ChatGPT/Codex, only for language polishing and clarity improvement. The authors reviewed all AI-assisted edits and take full responsibility for the manuscript.
\end{acks}

\bibliographystyle{ACM-Reference-Format}
\bibliography{references}

\section*{Appendix A: Deployment Feasibility Note}

RAPID is intended as an early-stage decision-support pipeline, so deployment feasibility depends on API cost, latency, and throughput as much as model accuracy. We estimate cost from the current code structure: DPA runs three OpenAI calls for mode recognition, image-grounded reasoning, and task planning; IRA first applies local IQA screening and then conditionally calls Gemini for restoration planning or image generation; DRA runs one Gemini call per image pair; and DReA runs one LLM call over structured visual evidence. For case \(i\), the approximate API cost is
\begin{equation}
C_i=\sum_{a\in\mathcal{A}}
\left(
\frac{p^{\mathrm{in}}_{m(a)}u_{i,a}}{10^6}
+\frac{p^{\mathrm{out}}_{m(a)}v_{i,a}}{10^6}
+p^{\mathrm{img}}_{m(a)}q_{i,a}
\right),
\label{eq:deployment-cost}
\end{equation}
where \(a\) indexes active agents, \(m(a)\) is the selected model, \(u_{i,a}\) and \(v_{i,a}\) are input and output token counts, and \(q_{i,a}\) counts image-generation or image-editing calls. Public prices change over time. As of May 17, 2026, OpenAI lists GPT-5.4 mini at \$0.75 per million input tokens, \$4.50 per million output tokens, and \$0.075 per million cached input tokens \cite{OpenAI2026Pricing}. Google lists Gemini 2.5 Flash at \$0.30 input and \$2.50 output per million tokens, Gemini 2.5 Pro at \$1.25 input and \$10.00 output for prompts up to 200k tokens, Gemini 3.1 Pro Preview at \$2.00 input and \$12.00 output for prompts up to 200k tokens, and Gemini 3 Pro Image Preview at about \$0.0011 per image input and \$0.134 per 1K/2K output image \cite{Google2026GeminiPricing}. The table below uses standard processing prices, one or two 1024-pixel input images per case, prompt sizes observed in the repository scripts, and moderate JSON or paragraph-length outputs. It excludes local CPU/GPU time, storage, free-tier credits, retries, and manual review.

\begin{samepage}
\noindent\textbf{Table A1. Per-task deployment planning estimates.}
\begin{center}
\scriptsize
\setlength{\tabcolsep}{2.5pt}
\renewcommand{\arraystretch}{1.12}
\begin{tabularx}{\columnwidth}{>{\raggedright\arraybackslash}p{0.22\columnwidth}>{\raggedright\arraybackslash}p{0.23\columnwidth}>{\raggedright\arraybackslash}p{0.23\columnwidth}>{\raggedright\arraybackslash}X}
\toprule
\textbf{Task} & \textbf{Typical calls} & \textbf{Estimated cost per case} & \textbf{Estimated response time} \\
\midrule
DPA mode and hazard routing & 1 OpenAI mini vision call & \$0.002--\$0.006 & 2--6 s \\
DPA image-grounded reasoning & 1 OpenAI mini vision call & \$0.003--\$0.010 & 4--12 s \\
DPA task planning & 1 OpenAI mini text call & \(<\)\$0.002 & 1--4 s \\
IRA local quality screening & Local IQA and rule checks only & \$0 & \(<\)1--2 s per image \\
IRA planner restoration & 1 Gemini 2.5 Flash image-text call, only if restoration is needed & \$0.001--\$0.003 per image & 1--5 s per image \\
IRA generative restoration & 1 Gemini 3 Pro Image Preview edit call, optional & \$0.135--\$0.245 per restored image & 8--30 s per image \\
DRA cross-view severity/object recognition & 1 Gemini 2.5 Pro image-text call for SVI+RSI & \$0.006--\$0.015 & 6--18 s plus the scripted 1 s batch pause \\
DRA bi-temporal severity/object recognition & 1 Gemini 3 Pro-class image-text call for pre/post SVI & \$0.010--\$0.025 & 8--25 s plus the scripted 1 s batch pause \\
DReA report synthesis & 1 Gemini Flash/Pro text call over structured evidence & \$0.003--\$0.013 & 3--10 s \\
\bottomrule
\end{tabularx}
\end{center}
\end{samepage}

Serial latency is \(T_i \approx T_{\mathrm{DPA}}+\mathbb{1}_{\mathrm{IRA}}T_{\mathrm{IRA}}+T_{\mathrm{DRA}}+T_{\mathrm{DReA}}\), where \(\mathbb{1}_{\mathrm{IRA}}\) indicates whether restoration is invoked. A typical no-restoration case therefore costs roughly \$0.02--\$0.06 and takes about 20--55 seconds in serial execution; one generative restoration adds about \$0.135--\$0.245 and 8--30 seconds. For the 50-pair evaluation scale used in the current scripts, the no-restoration API cost is approximately \$1--\$3, while editing one image for every case would add about \$6.75--\$12.25. Parallel model calls reduce wall-clock time, and DReA can be skipped for triage-only runs. Because rate limits vary by provider, model, request type, and usage tier \cite{OpenAI2026RateLimits,Google2026GeminiRateLimits}, deployments should log tokens, time, retries, and throttling.

\end{document}